\documentclass[10pt,twocolumn,letterpaper]{article}

\usepackage{iccv}              
\usepackage{bbding}
\usepackage{float} 
\usepackage[accsupp]{axessibility}  
\usepackage{fontawesome5}

\usepackage{algorithm} 
\usepackage{algorithmic}
\usepackage{multirow}

\definecolor{iccvblue}{rgb}{0.21,0.49,0.74}
\usepackage[pagebackref,breaklinks,colorlinks,allcolors=iccvblue]{hyperref}

\def\paperID{6345} 
\def\confName{ICCV}
\def\confYear{2025}

\title{

Describe, Don't Dictate: Semantic Image Editing with Natural Language Intent

}

\author{\\
En Ci\textsuperscript{1,2$\star$}\quad 
Shanyan Guan\textsuperscript{2$\star$}\quad 
Yanhao Ge\textsuperscript{2}\quad 
Yilin Zhang\textsuperscript{1}\quad  \\
Wei Li\textsuperscript{2}\quad 
Zhenyu Zhang\textsuperscript{1}\quad 
Jian Yang\textsuperscript{1}\quad 
Ying Tai\textsuperscript{1}\textsuperscript{\faEnvelope }\quad \\
\textsuperscript{1}Nanjing University, China\quad
\textsuperscript{2}vivo, China \\
{\tt\small \{cien,guanshanyan,halege\}@vivo.com, \{zhenyuzhang, yingtai\}@nju.edu.cn} 
}

\begin{document}
\newcommand\blfootnote[1]{%
  \begingroup
  \renewcommand\thefootnote{}\footnote{#1}%
  \addtocounter{footnote}{-1}%
  \endgroup
}

\twocolumn[{%
	\renewcommand\twocolumn[1][]{#1}%
	\maketitle
        \vspace{-12mm}
	\begin{center}
        \centering
        \includegraphics[width=\textwidth]{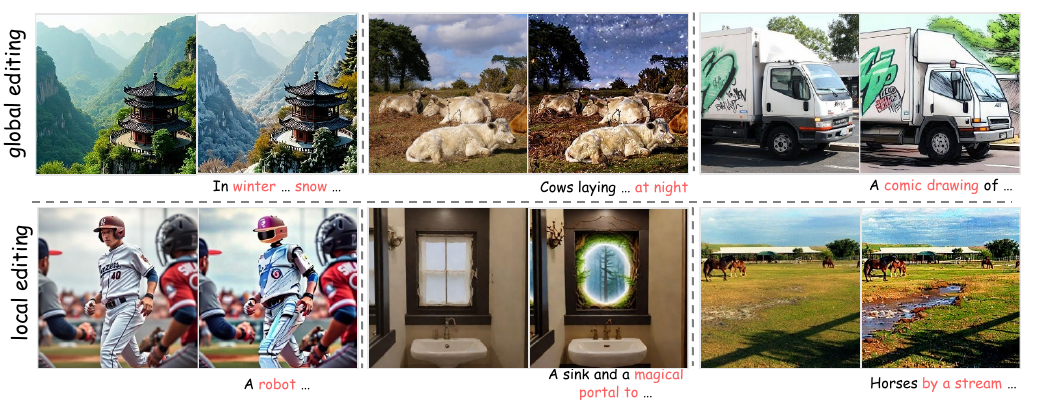}
        \vspace{-8mm}
        \captionof{figure}{Our \textbf{DescriptiveEdit} enables description-based rather than instruction-based image editing, achieving strong performance in both global editing (Top) and local editing (Bottom). 
        The original image is on the left, with the edit description below each edited image.} 
        \label{fig:head}
    \end{center}
}]

\blfootnote{$\star$~Equal contributions.\quad \textsuperscript{\faEnvelope}Correspondence to: Ying Tai.}

\begin{abstract}

Despite the progress in text-to-image generation, semantic image editing remains a challenge. 
Inversion-based algorithms unavoidably introduce reconstruction errors, while instruction-based models mainly suffer from limited dataset quality and scale.  
To address these problems, we propose a descriptive-prompt-based editing framework, named DescriptiveEdit.
The core idea is to re-frame `instruction-based image editing' as `reference-image-based text-to-image generation', which preserves the generative power of well-trained Text-to-Image models without architectural modifications or inversion. 
Specifically, taking the reference image and a prompt as input, we introduce a Cross-Attentive UNet, which newly adds attention bridges to inject reference image features into the prompt-to-edit-image generation process. 
Owing to its text-to-image nature, DescriptiveEdit overcomes limitations in instruction dataset quality, integrates seamlessly with ControlNet, IP-Adapter, and other extensions, and is more scalable.
Experiments on the Emu Edit benchmark show it improves editing accuracy and consistency.

\end{abstract}

\section{Introduction}
\label{sec:intro}

Text-to-image (T2I) generation—\textit{synthesizing images from textual prompts describing desired content}—has made significant breakthroughs~\cite{flux,esser2024scaling,chen2023pixart,sdxl}, driven by advances in three key areas: large high-quality datasets
(\eg, LAION-5B~\cite{schuhmann2022laion}), efficient and scalable network architectures (\eg, Latent Diffusion Models~\cite{rombach2022high}, Diffusion Transformers~\cite{peebles2023scalable}), and theoretical progress in modeling high-dimensional data distributions~\cite{lipman2022flow,li2024autoregressive,ho2020denoising,tian2024visual}.
%
However, semantic image editing\textemdash\textit{``generating'' what people want to change on the given images (still is text-guided)}\textemdash remains fundamentally challenging. 


Inversion-based techniques~\cite{p2p,pnp,postedit,rfinversion,dit4edit,rfedit,kvedit} invert the input image to a noisy latent and regenerate a new image guided by an edit prompt (\eg, ``a smiling person'').
However, the inverting process suffers from reconstruction errors and inefficiency due to its iterative optimization. 
Recently, modifying the T2I model architecture and training it to adhere to specific editing instructions (\eg, ``make the person smile'') yields harmonious and accurate editing results~\cite{smartedit,ultraedit,mgie,anyedit,instructpix2pix}. 
However, there are two limitations we need to consider.
First is the data bottleneck. 
Collecting a large-scale dataset of instructional training data—comprised of input images, edited images, and edit instructions—is both challenging and costly.
Compared to popular T2I generation dataset (\eg, LAION-5B), existing instruction-based datasets, such as AnyEdit~\cite{anyedit} ($\sim$2.5M) and UltraEdit~\cite{ultraedit} ($\sim$4M) are significantly smaller and less diverse.
Second, modifying the architecture entails substantial computational costs for retraining the diffusion model and can hinder compatibility with diverse pretrained T2I models and their associated ecosystem tools (\eg, ControlNet~\cite{Zhangcontrolnet}, IP-Adapter~\cite{ipadapter}).

These challenges raise a key question: Can we ``generate'' the desired editing intent directly from descriptive prompts instead of relying on instructions? 
If so, text-to-image generation and editing would merge into a unified computational framework, inherently solving the issues of training data scale and compatibility.
Moreover, the descriptive prompt contains more details, enabling more precise edits.
The underlying motivation is that instruction-based editing can be equivalently understood as a two-stage process: ``instructions $\rightarrow$ edit descriptions $\rightarrow$ edit images''. 
%


To answer this question, we propose DescriptiveEdit, a novel image editing framework that reinterprets the role of text guidance by transitioning from instruction-based to description-driven editing. Instead of modifying diffusion model architectures or requiring additional training on instruction-labeled datasets, DescriptiveEdit leverages pre-trained T2I models in a plug-and-play manner, preserving their original generative capacity.
%
%
To achieve this, we introduce a Cross-Attentive UNet, which injects reference image features into the denoising process via a lightweight attention bridge (newly added attention layers). This mechanism enables the model to align the edited output with the structural and semantic cues of the original image, while ensuring faithful adherence to the descriptive prompt. Moreover, instead of modifying the model's core architecture, we apply low-rank parameter tuning (LoRA)~\cite{lora}, enabling parameter-efficient adaptation without disrupting the pre-trained generative capabilities.

Furthermore, DescriptiveEdit seamlessly integrates with existing T2I pipelines without requiring model retraining. Unlike instruction-based methods such as IP2P~\cite{instructpix2pix}, which fine-tune the full UNet ($\sim$860M parameters), our approach introduces only a lightweight set of trainable parameters ($\sim$75M) while maintaining superior editing accuracy and structural consistency. Additionally, DescriptiveEdit remains fully compatible with community-driven extensions, such as ControlNet~\cite{Zhangcontrolnet} and IP-Adapter~\cite{ipadapter}, allowing enhanced control over generation while preserving the flexibility of pre-trained models.

In summary, our contributions are as follows:
\begin{itemize}
    \item We re-examine the role of T2I models in image editing and propose a description-based semantic editing paradigm that preserves generative quality.
    \item We present a non-invasive adaptation framework that merges attention-based feature projection with parameter-efficient tuning, providing both flexibility and reusability.
    \item Extensive experiments demonstrate our method’s superior performance in editing accuracy and image consistency compared to existing approaches.
\end{itemize}

\section{Related Work}
\label{sec:related}

\paragraph{Text-to-Image Generation.}
%
Modern text-to-image synthesis~\cite{gan,dalle1,glide,dalle2,sd,sd3,flux} has undergone three evolutionary phases. Early GAN-based approaches~\cite{gan} struggled with mode collapse and limited diversity. Inspired by nonequilibrium thermodynamics~\cite{sohl2015deep}, the diffusion revolution began with GLIDE~\cite{glide}, which established pixel-space text-image alignment through iterative denoising. Latent diffusion models (LDM)~\cite{sd} later improved efficiency by operating in compressed feature space while maintaining fidelity. Recent architectural innovations like Diffusion Transformers (DiT)~\cite{dit} replace convolutional UNets with pure attention mechanisms, enabling superior global coherence and scalability. Parallel developments in flow-based models~\cite{flux} offer alternative probabilistic mappings from noise to image space. These advancements collectively establish the foundation for semantic image editing.

\vspace{-4mm}
\paragraph{Semantic Image Editing.}
Semantic image editing demands precise modifications while preserving fidelity, making it more challenging than T2I generation. Existing methods fall into training-based and training-free approaches.
Training-based methods improve control but often degrade generative quality. Some fine-tune entire models~\cite{imageneditor,instructpix2pix,magicbrush,ultraedit}, causing artifacts, while others use MLLMs\cite{mgie,smartedit} or MoE architectures\cite{emuedit,anyedit,omniedit}, which enhance task-specific performance but require costly retraining.
Training-free methods exploit pre-trained models but struggle with efficiency and accuracy. Early works modified noise levels\cite{blendeddiffusion}, later methods adjusted attention mechanisms\cite{p2p,pnp,masactrl,towards}, and Flux-based models~\cite{rfedit,rfinversion,dit4edit,RAGD} improved sampling efficiency. However, inversion reliance increases computational cost.
We introduces an efficient training paradigm that preserves generative ability while enabling general-purpose editing via descriptive text, achieving a better trade-off between control, efficiency, and generalization.

\begin{figure*}[t]
  \centering
  \includegraphics[width=0.96\textwidth]{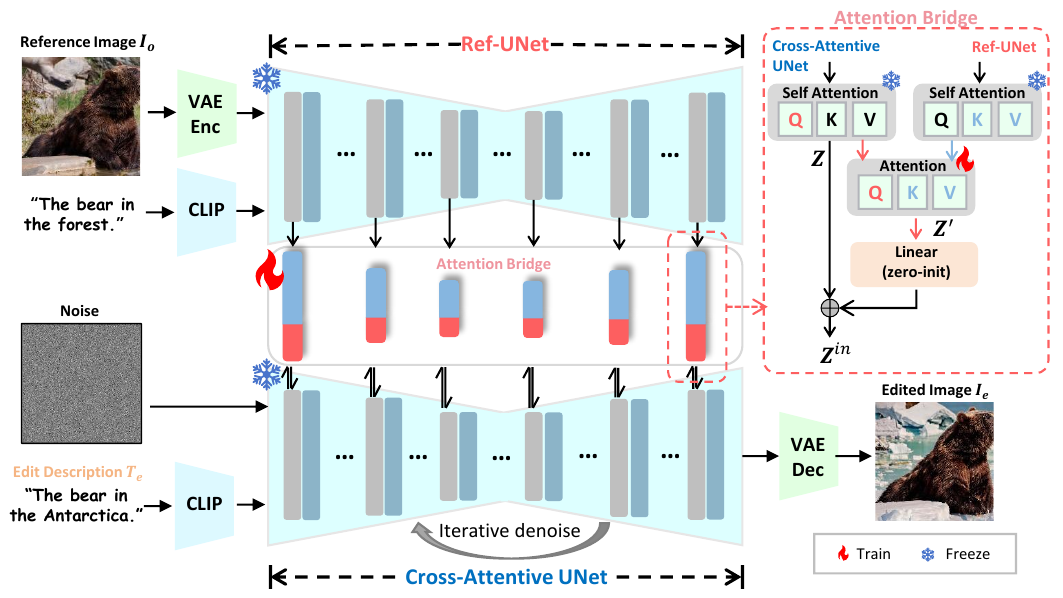}
  \vspace{-3mm}
    \caption{Overview of the \textbf{DescriptiveEdit} framework. 
    We introduce attention-bridge layers between two identical UNets, enabling the denoising UNet to function as a Cross-Attentive UNet.
    }
    \vspace{-5mm}
  \label{fig:framework}
\end{figure*}

\section{DescriptiveEdit}
\label{sec:method}

In this work, we propose, DescriptiveEdit, a novel approach to text-guided semantic image editing by describing the final desired effect rather than providing direct editing instructions as in conventional instruction-based methods. 
In particular, DescriptiveEdit leverages pre-trained T2I models as a fixed foundational component without altering their weights, thereby avoiding any modifications to the architecture or additional training processes.

The remainder of the paper is organized as follows. In Sec.~\ref{sec:problem_formulation}, we formally define the problem. Sec.~\ref{sec:framwork_archi} details the framework and its key components. Finally, Sec.~\ref{sec:traininginference} presents the procedures for training and inference.

\subsection{Problem Formulation}
\label{sec:problem_formulation}
{Instruction-based image editing works} formulate the semantic image editing task as follows: given an original image and the instructions describing the desired modifications (\eg add firework to the sky), a model is trained to generate the corresponding edited output. 
In our approach, we reframe this process into two distinct stages: ``\text{instructions} $\rightarrow$ \text{edit descriptions} $\rightarrow$ \text{edited images}.''
Here, the \emph{edit description} (\eg, fireworks are blooming in the sky) can be either user-specified or automatically derived from the original image and instructions using large visual language models (\eg, QWen-VL~\cite{wang2024qwen2}).
The unexplored challenge, therefore, is to design a model that can effectively translate an edit description into the corresponding edited image.

Formally, given the edit description $\boldsymbol{T}_{e}$ and the original image $\boldsymbol{I}_{o}$, the formulation of our DescriptiveEdit is:
\begin{align}
    \boldsymbol{I}_{e} \leftarrow \mathcal{F}_{\boldsymbol{\theta}} (\boldsymbol{T}_{e}|\boldsymbol{I}_{o}),
\end{align}
where $\boldsymbol{I}_{e}$ is the edited image, $\mathcal{F}_{\boldsymbol{\theta}}$ represents the editing model, and $\boldsymbol{\theta}$ is its parameters. $\boldsymbol{T}_{e}$ actually is the ``prompt'' in context of T2I models, and thus $\mathcal{F}_{\boldsymbol{\theta}}$ is essentially a T2I diffusion model conditioned on the original image $\boldsymbol{I}_{o}$.

But, since the T2I model does not support image-based conditioning, we seek adjustments. Unlike InstructPix2Pix, which modifies structure and input at the cost of compatibility and training efficiency, we aim for a lightweight, flexible solution that conditions on the original image without altering the model’s architecture or pretrained weights.

\subsection{Framework Architecture}
\label{sec:framwork_archi}

As illustrated in Fig.~\ref{fig:framework}, we introduce an attention fusion mechanism that operates within the latent space of the frozen model, enabling effective integration of original image $\boldsymbol{I}_{o}$ information into the editing process (\ie, the edit description $\boldsymbol{T}_{e}$ $\rightarrow$ the edited image $\boldsymbol{I}_{e}$). To ensure efficiency, we adopt low-rank parameter updates using LoRA~\cite{lora}.


\subsubsection{Cross-Attentive UNet with Attention Bridge}
\label{sec:attention_fusion}
LDMs commonly employ two conditioning schemes: cross-attention (\eg, Stable Diffusion~\cite{sd}, IP-Adaptor~\cite{ipadapter}) and a separate UNet for feature extraction (\eg, ControlNet, Animate-Anyone~\cite{hu2024animate}), each with its own strengths and limitations.  
Cross-attention (CA) is a lightweight conditioning mechanism with minimal training overhead. It integrates conditional information $\boldsymbol{c}$ via the general formulation:  
\begin{align}
    \boldsymbol{Z}^{\prime} = \text{CA}(\boldsymbol{Q}, \boldsymbol{K}_{\boldsymbol{c}}, \boldsymbol{V}_{\boldsymbol{c}}) = \text{softmax} \left(\frac{\boldsymbol{Q} \boldsymbol{K}^{\mathsf{T}}_{\boldsymbol{c}}}{\sqrt{d}}\right) \cdot \boldsymbol{V}_{\boldsymbol{c}},  
    \label{eq:ca}
\end{align}  
where  
\begin{align}
    \boldsymbol{Q} = \boldsymbol{Z} \boldsymbol{W}_{q}, \quad  
    \boldsymbol{K}_{\boldsymbol{c}} = \boldsymbol{f}_{\boldsymbol{c}} \boldsymbol{W}_{k}, \quad  
    \boldsymbol{V}_{\boldsymbol{c}} = \boldsymbol{f}_{\boldsymbol{c}} \boldsymbol{W}_{v}.
\end{align}  
Here, $\boldsymbol{Z}$ is the intermediate representation of the UNet, $\boldsymbol{f}_{\boldsymbol{c}}$ represents the feature encoding of the conditioning input $\boldsymbol{c}$, $\boldsymbol{W}_{q}$, $\boldsymbol{W}_{k}$, and $\boldsymbol{W}_{v}$ are learnable projection matrices.  
Although CA-based conditioning is computationally efficient, the input feature $\boldsymbol{f}_{\boldsymbol{c}}$ primarily encodes high-level semantics, enforcing only semantic relevance to the condition $\boldsymbol{c}$ (as also discussed in Animate-Anyone).  
For low-level alignment, an alternative approach involves using a separate UNet (a copy of the denoising UNet) to extract features from the conditional image. These features are then fused with the denoising UNet via channel-wise concatenation (refer to ControlNet). However, this modification alters the input feature dimensions, requiring fine-tuning on large-scale datasets, which significantly increases training costs.

To seamlessly integrate the strengths of both conditioning schemes—cross-attention and separate UNet encoding—we propose a novel image-conditioning algorithm for semantic image editing.  
To enable low-level alignment the edit output and the input image, we still use a separate UNet (so-called Ref-UNet, inherent weights from the denoising UNet) to encode the input images. 
At the self-attention layers of both UNets~\footnote{It's a common practice since self-attention dominates spatial information~\cite{p2p,glide,kumari2023multi}.}, we perform cross-attention by taking the key ($\boldsymbol{K}_{\boldsymbol{T}_{e}}$) and value ($\boldsymbol{V}_{\boldsymbol{T}_{e}}$) from the denoising UNet’s self-attention and the query ($\boldsymbol{Q}_{\boldsymbol{I}_{o}}$) from the Ref-UNet’s self-attention (see Eq.~\ref{eq:ca}).  
The formulation is:
\begin{align}
    \boldsymbol{Z}^{\prime} = \text{CA}(\boldsymbol{Q}_{\boldsymbol{I}_{o}}, \boldsymbol{K}_{\boldsymbol{T}_{e}}, \boldsymbol{V}_{\boldsymbol{T}_{e}}).
\end{align}
$\boldsymbol{Z}^{\prime}$ is then added to $\boldsymbol{Z}$ (the self-attention output of the denoising UNet ).
The newly introduced cross-attention mechanism acts as a bridge, enabling the denoising UNet to incorporate information from the Ref-UNet, effectively forming a Cross-Attentive UNet.

Our proposed cross-attentive denoising scheme requires training only the newly added CA layers, while all other parameters remain frozen.  
Moreover, it is important to note that cross-attentive denoising is not restricted to UNet-based architectures; it is also compatible with DiT-based architectures. By replacing UNet with DiT and applying the cross-attention mechanism used in DiT, the same framework can be extended. Refer to Fig.~\ref{fig:flux} for qualitative results.


\subsubsection{Balancing Reference and Editing}
\label{sec:balance}
A straightforward approach to incorporating reference features is to directly add the self-attention output of the denoising UNet, $\boldsymbol{Z}$, with the reference-enhanced feature, $\boldsymbol{Z}^{\prime}$. 
%
However, this manner often overpowers the original features and nullifies the intended editing effect. Further analysis is provided in Appendix.

To address this, we add a learnable linear mapping to establish a \textit{dynamic equilibrium} between these components:
\begin{align}
    \boldsymbol{Z}^{\text{in}} = \boldsymbol{Z} + \text{Linear}(\boldsymbol{Z}^{\prime}),
\end{align}
where $\text{Linear}(\cdot)$ is a learnable transformation initialized with zero weights. This design offers two key advantages: (1) it adaptively balances the generative prior ($\boldsymbol{Z}$) and reference guidance ($\boldsymbol{Z}^{\prime}$) by learning optimal contribution weights; (2) zero initialization ensures $\text{Linear}(\boldsymbol{Z}^{\prime}) \approx 0$ at the start of training, preserving the base model’s behavior while gradually integrating reference features.

\subsection{Training and Inference}
\label{sec:traininginference}
\paragraph{Training.}
Inspired by classifier-free guidance~\cite{classifier}, we randomly set edit descriptions and original images empty for 5\% of examples independently during training. Both the edited image $\boldsymbol{Z}_{e}$ and the original image $\boldsymbol{Z}_{o}$ are simultaneously fed into the model. But, if the same noise is applied to them, crucial reference information may be lost, hindering the model's ability to learn the editing capability. Inspired by Diffusion Forcing~\cite{diffusionforce}, where the model is trained with inputs containing independent noise levels, we only apply noise to the edited image. 
For parameter-efficient adaptation, we initialize the attention bridge layers by copying the weights from pretrained self-attention layers and fine-tune them using LoRA~\cite{lora}. Meanwhile, the newly introduced learnable mapping layers follow ControlNet~\cite{Zhangcontrolnet}'s zero-initialization strategy to preserve the original model behavior at initialization.
Finally, we minimize the following latent diffusion objective:
\begin{align}
    \mathcal{L} = \mathbb{E}_{\boldsymbol{Z}_{e}^{0}, \boldsymbol{Z}_{o}^{0}, \boldsymbol{\epsilon}, t, s} \left[ \|\boldsymbol{\epsilon} - \boldsymbol{\epsilon}_\theta(\boldsymbol{Z}_{e}^{t}, t, \boldsymbol{T}_{e}, \boldsymbol{Z}_{o}^{s}, s)\|^2 \right],
\end{align}
where $t$  is the diffusion timestep for the edited image, $s$ is the fixed timestep for the original image set to $s=0$, $ \boldsymbol{Z}_{e}^{t} $ denotes the noisy edited image latent representation at timestep $t$ and $\boldsymbol{Z}_{o}^{s} $  represents the clean original image latent representation at timestep $s$.

\vspace{-4mm}
\paragraph{Inference.}
Since our training process incorporates conditional information from the original image in addition to the edited image and edit description, we redefine the denoising formulation, drawing inspiration from IP2P~\cite{instructpix2pix}. The modified denoising process is formulated as follows:
\begin{equation}
    \begin{aligned}
        \tilde{\boldsymbol{\epsilon}}_{\theta}(\boldsymbol{Z}_{e}^{t}, \boldsymbol{Z}_{o}^{s}, \boldsymbol{T}_{e}) &=  \boldsymbol{\epsilon}_{\theta}(\boldsymbol{Z}_{e}^{t}, \varnothing, \varnothing) \\
        &+ \lambda_I \cdot \big( \boldsymbol{\epsilon}_{\theta}(\boldsymbol{Z}_{e}^{t}, \boldsymbol{Z}_{o}^{s}, \varnothing) - \boldsymbol{\epsilon}_{\theta} (\boldsymbol{Z}_{e}^{t}, \varnothing, \varnothing) \big) \\
        &+ \lambda_T \cdot \big( \boldsymbol{\epsilon}_{\theta}(\boldsymbol{Z}_{e}^{t}, \boldsymbol{Z}_{o}^{s}, \boldsymbol{T}_{e}) - \boldsymbol{\epsilon}_{\theta}(\boldsymbol{Z}_{e}^{t}, \boldsymbol{Z}_{o}^{s}, \varnothing) \big),
    \end{aligned}
\end{equation}
The terms $ \lambda_I $ and $ \lambda_T $ control the guidance strength for the original image and textual conditions, respectively. This formulation ensures that the model effectively integrates both textual and original image conditions while maintaining its generative capabilities.

\section{Experiments}
\label{sec:experiments}


\subsection{Experimental Setup}
\label{setup}

\paragraph{Training Details.} To ensure fair comparisons, we adopt \texttt{Stable Diffusion v1.5}~\cite{sd} as the backbone, aligning with most training-based methods~\cite{instructpix2pix,magicbrush,anyedit}. The model is optimized using the AdamW~\cite{adamw} optimizer with a learning rate of \( 1\times10^{-5} \). For parameter-efficient fine-tuning, we apply LoRA~\cite{lora} to the attention bridge layers, setting the rank to 64 and $\alpha$ to 64. Additional training protocols are detailed in Appendix~A.

\vspace{-4mm}
\paragraph{Datasets.} Our model is trained on the UltraEdit dataset~\cite{ultraedit}, which contains approximately 4M text-image pairs. For evaluation, we use the Emu Edit Test benchmark~\cite{emuedit} following~\cite{emuedit,anyedit,ultraedit}. 
We found inconsistencies in this benchmark's quality, where identical captions (e.g., “a train station in a city”) are assigned to both source and target images, regardless of their actual visual content.
%
%
To ensure a fair and sound evaluation, we manually filter out non-compliant samples before calculating the metrics.


\vspace{-4mm}
\paragraph{Baselines.} 

We compare our approach against representative baselines from two categories:  
{(1)~\textit{Training-free} methods, including MasaCtrl~\cite{masactrl}, RF-Edit~\cite{rfedit},PnPInversion~\cite{pnpinversion}, FPE~\cite{FPE}, and Turboedit~\cite{turboedit}. 
(2)~\textit{Training-based} approaches, including InstructPix2Pix~\cite{instructpix2pix}, MagicBrush~\cite{magicbrush}, EmuEdit~\cite{emuedit}, AnyEdit~\cite{anyedit}, and BrushEdit~\cite{li2024brushedit}. Comprehensive baseline specifications are provided in Appendix~B.

\vspace{-4mm}
\paragraph{Evaluation Metrics.} We evaluate the performance of semantic image editing models from three perspectives: 
(1) \textit{Instruction Adherence}: We use the CLIP-T score~\cite{clip} to quantify the alignment between the edit description and the edited image. 
(2) \textit{Image Consistency}: Pixel-level preservation is measured using L1 and L2 distances between original and edited images. Semantic-level consistency is assessed using CLIP-I~\cite{clip} and DINO-I~\cite{dino} feature similarities, while structural integrity is evaluated via the Structural Similarity Index (SSIM)~\cite{ssim}. Additionally, we adopt the Learned Perceptual Image Patch Similarity (LPIPS)~\cite{lpips} to evaluate perceptual consistency.
(3) \textit{Image Quality}: We employ the Peak Signal-to-Noise Ratio (PSNR) to assess pixel-wise reconstruction accuracy. 
Comprehensive metric implementations are provided in Appendix~C.

\begin{table*}[t]
    \centering
    \caption{
    Quantitative Comparison on the EMU-Edit test-set. 
    We evaluate models in terms of instruction adherence (CLIP-T), image consistency (L1, L2, SSIM, DINO-I, CLIP-I, LPIPS), and image quality (PSNR). 
    $\ast$ indicates results from the original paper due to unavailable code. 
    $\dagger$ indicates results obtained by retraining models on our training dataset.
    The best and second-best results are in \textbf{bold} and \underline{underlined}. 
    %
    }
    \label{tab:comparison}
    \resizebox{0.9\textwidth}{!}{%
    \begin{tabular}{l|c|cccccccc}
        \toprule
        Model & Training-based? & L1 $\downarrow$ & L2 $\downarrow$ & LPIPS $\downarrow$ & PSNR $\uparrow$ & SSIM $\uparrow$ & DINO-I $\uparrow$ & CLIP-I $\uparrow$ & CLIP-T $\uparrow$ \\
        \midrule
        MasaCtrl~\cite{masactrl} & \XSolidBrush & 0.072 & 0.014 & 0.174 & 19.31 & 0.654 & 0.797 & \underline{0.863} & 0.299 \\
        RF-Edit~\cite{rfedit}  & \XSolidBrush & 0.096  & 0.022  & 0.317  & 17.10  & 0.554  & 0.553  & 0.757  & \textbf{0.319} \\
        PnPInversion~\cite{pnpinversion} & \XSolidBrush & 0.084 & 0.014 & 0.205 & 18.92 & 0.639 & 0.749 & 0.802 & 0.307 \\ 
        FPE~\cite{FPE} & \XSolidBrush & 0.071 & \underline{0.012} & 0.185 & \underline{20.41} & \textbf{0.664} & 0.714 & 0.822 & 0.293 \\
        TurboEdit~\cite{turboedit}  & \XSolidBrush & 0.095 & 0.023 & 0.234 & 17.56 & 0.598 & 0.659 & 0.787 & 0.313 \\ 
        \midrule
        IP2P$^{\dagger}$~\cite{instructpix2pix} & \Checkmark & 0.083 & 0.015 & 0.210 & 20.03 & 0.619 & 0.740 & 0.805 & 0.293 \\ 
        MagicBrush~\cite{magicbrush} & \Checkmark & 0.082  & 0.026  & 0.179  & 19.48  & 0.655  & 0.752  & 0.837  & 0.305 \\
        EmuEdit*~\cite{emuedit} & \Checkmark & 0.094  & -      & -      & -      & -      & \underline{0.819}  & 0.859  & 0.231 \\
        AnyEdit$^{\dagger}$~\cite{anyedit} & \Checkmark & \underline{0.067} & 0.020 & \underline{0.147} & 19.81 & 0.657 & 0.809 & 0.832 & 0.271 \\ 
        BrushEdit~\cite{li2024brushedit} & \Checkmark & 0.143  & 0.050  & 0.272  & 13.51  & 0.451  & 0.601  & 0.782  & 0.299 \\
        \midrule
        \textbf{DescribeEdit(Ours)}  & \Checkmark & \textbf{0.065}  & \textbf{0.011}  & \textbf{0.139}  & \textbf{20.99}  & \underline{0.661}  & \textbf{0.843}  & \textbf{0.874}  & \underline{0.315} \\
        \bottomrule
    \end{tabular}
    }
    \vspace{-3mm}
\end{table*}

\subsection{Quantitative Comparison}
\label{quantitative comparison}

The quantitative results are reported in Tab.~\ref{tab:comparison}.
We compared our method with both training-free and training-based baselines across multiple evaluation metrics.
%
%
%
In terms of \textit{image consistency}, our method exhibits the lowest L1 (0.065) and L2 (0.011) distances, ensuring minimal pixel-level deviation from the reference. Moreover, it achieves the highest DINO-I (0.843) and CLIP-I (0.874) scores, demonstrating superior semantic and feature-level consistency. 
%
Additionally, our model obtains the best LPIPS score (0.139) and a competitive SSIM score (0.661), highlighting its strong perceptual similarity and structural coherence with the reference image. The slightly higher SSIM of FPE likely stems from its focus on preserving spatial structure, which reduces flexibility for non-rigid edits.
%
Regarding \textit{instruction adherence}, our model achieves a competitive CLIP-T score (0.315), which is comparable to RF-Edit (0.319), indicating strong alignment between generated images and textual instructions. 
%
For \textit{image quality}, our model outperforms all baselines with the highest PSNR (20.99), indicating superior pixel-wise reconstruction accuracy. The slightly lower CLIP-T score compared to RF-Edit likely results from the inherent characteristics of training-free methods, which prioritize instruction adherence at the expense of fine-grained visual consistency. Overall, our results suggest that training-based methods like ours can achieve a more optimal trade-off, ensuring high-fidelity, description-driven edits while maintaining strong consistency with reference images. This highlights the practical utility of our approach in real-world semantic image editing tasks.
\begin{figure}[t]
    \centering
    \includegraphics[width=\linewidth]{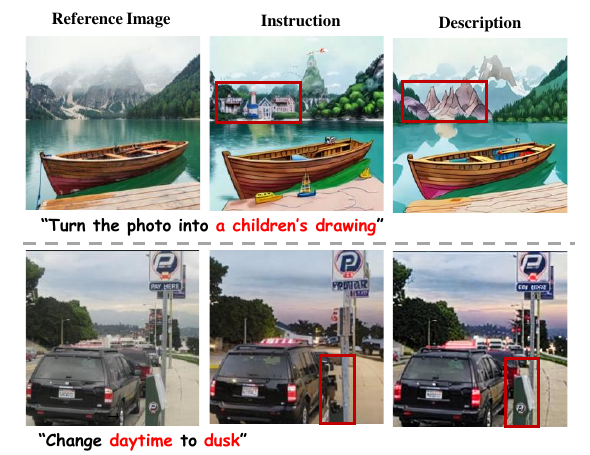}
    \vspace{-6mm}
    \caption{Qualitative comparison between instruction-based and description-based editing. Description prompts yield richer, more precise edits in terms of structure and semantics.}
    \vspace{-6mm}
    \label{fig:qual_instruction_vs_description}
\end{figure}
\subsection{Qualitative Comparison}
\label{qualitative_comparison}
\begin{figure*}[t]
    \centering
    \includegraphics[width=\linewidth]{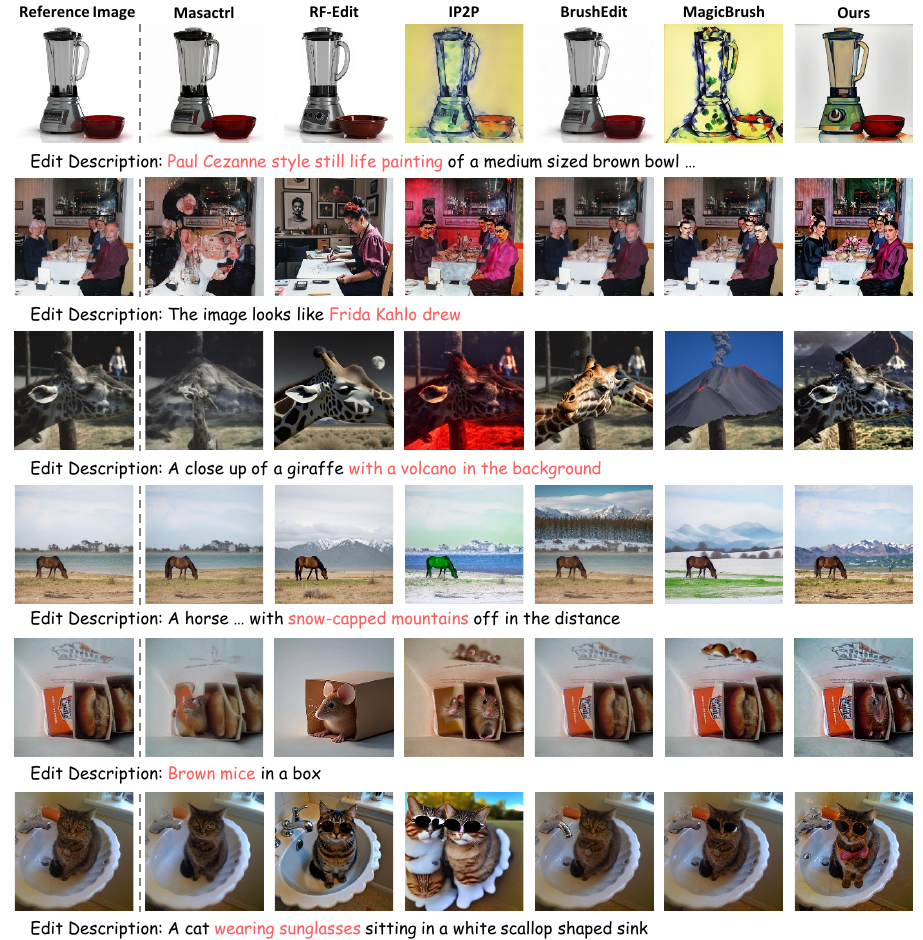}
    \vspace{-3mm}
    \caption{Qualitative comparison of global and local edits across training-based and training-free methods.}
    \vspace{-5mm}
    \label{fig:qualitative_comparison}
\end{figure*}


Fig.~\ref{fig:qualitative_comparison} shows the qualitative comparisons on global and local editing tasks.
Our method outperforms both training-based and training-free approaches, including the flux-based RF-Edit, by achieving high-fidelity edits with strong semantic consistency, while existing methods lack robustness, often failing drastically in certain cases.

For global editing, our model effectively transforms styles and replaces backgrounds while preserving key foreground elements. As shown in Fig.~\ref{fig:qualitative_comparison}, it successfully integrates new backgrounds, such as volcanoes or snow-covered mountains, without introducing artifacts—whereas existing methods often fail to modify the background correctly or disrupt structural integrity. In local editing, our approach seamlessly modifies objects—\eg, replacing bread with a mice or adding glasses to a cat—while maintaining coherence, a challenge for prior methods.

\subsection{Ablations}
\begin{table}[t]
    \centering
    \caption{Quantitative comparison of description-based and instruction-based inputs for fine-tuning T2I models. The description-based approach achieves higher performance, demonstrating better alignment with the pre-trained T2I priors.}
    \vspace{-2mm}
    \label{tab:desc_vs_instr}
    \resizebox{0.9\columnwidth}{!}{%
    \begin{tabular}{lcccc}
        \toprule
         Method & CLIP-T $\uparrow$ & DINO-I $\uparrow$ & SSIM $\uparrow$ & PSNR $\uparrow$ \\
        \midrule
         Description  & \textbf{0.284} & \textbf{0.741} & \textbf{0.562} & \textbf{18.309} \\
         Instruction  & 0.272 & 0.739 & 0.551 & 18.123 \\
        \bottomrule
    \end{tabular}%
    }
\end{table}
\begin{table}[t]
    \centering
    \vspace{-2mm}
    \caption{Ablation study on the attention fusion strategies.
    \vspace{-2mm}
    }
    \label{tab:ablation}
    \resizebox{0.9\columnwidth}{!}{%
    \begin{tabular}{lcccc}
        \toprule
        Method & {CLIP-T} $\uparrow$ & {DINO-I} $\uparrow$ & {SSIM} $\uparrow$ & {PSNR} $\uparrow$ \\
        \midrule
        Direct Replacement  & 0.3005 & 0.6690 & 0.4261 & 13.78 \\
        Direct Addition & 0.3052 & 0.7532 & 0.4970 & 14.77 \\
        Ours & \textbf{0.3162} & \textbf{0.7931} & \textbf{0.6153} & \textbf{18.58} \\
        \bottomrule
    \end{tabular}
    }
    \vspace{-6mm}
\end{table}
\paragraph{Description vs. Instruction.}

\begin{figure*}[htbp]
    \centering
    \includegraphics[width=0.9\linewidth]{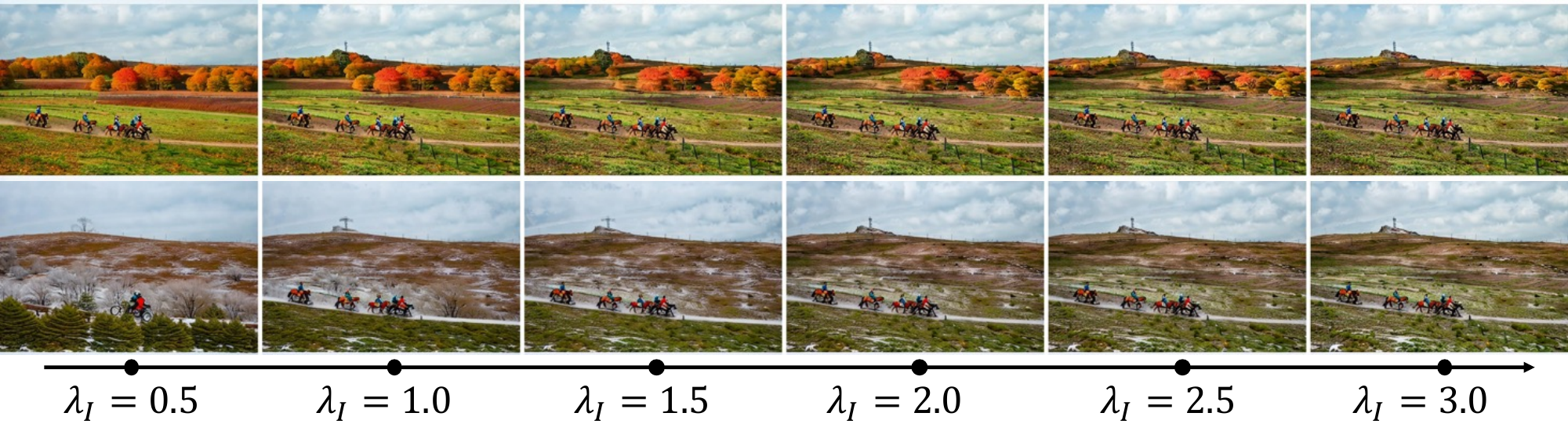}
    \vspace{-3mm}
    \caption{Effect of $\lambda_I$ on Image Editing (Top row: autumn; Bottom row: winter). From left to right, $\lambda_I$ increases linearly. }
    \label{fig:lambda_I_effect}
\end{figure*}
\begin{figure*}[h]
    \centering
    \includegraphics[width=0.9\linewidth]{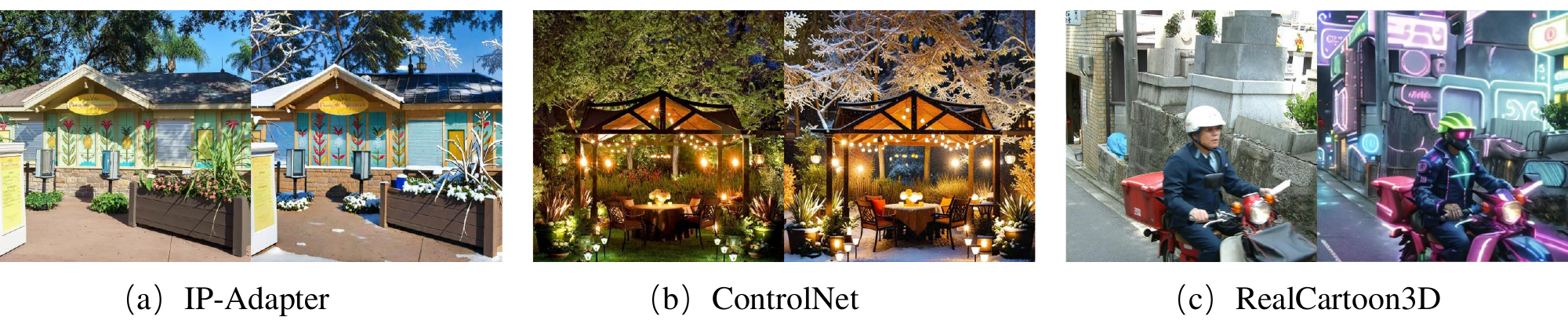}
    \vspace{-3mm}
    \caption{
    Cross-architecture compatibility demonstrated with: (a) IP-Adapter for seasonal transformation (original → winter), (b) ControlNet for seasonal transformation (original → winter), and (c) RealCartoon3D for cartoon stylization (original → 3D animated style).
    }
    \vspace{-5mm}
    \label{fig:community}
\end{figure*}
%

As discussed in Sec.\ref{sec:intro}, T2I models are pre-trained on large-scale description-image data, making descriptions more compatible than instructions for image editing. We conduct a experiment by varying only the text input type and observe that the description-based method consistently outperforms the instruction-based one across multiple metrics (Tab.\ref{tab:desc_vs_instr}). Beyond quantitative results, Fig.~\ref{fig:qual_instruction_vs_description} illustrates that description prompts lead to more accurate edits under the same intentions. This demonstrates that free-form descriptions offer better guidance by providing richer details and clearer intent, resulting in improved editing fidelity.

\vspace{-4mm}
\paragraph{Effect of $\lambda_I$ on Image Editing.}
To analyze the impact of $\lambda_I$, we conduct an ablation study by varying its intensity and observing the model's behavior. The results are visualized in Fig.~\ref{fig:lambda_I_effect}, illustrating seasonal transitions. When $\lambda_I$ is small, the model exhibits strong editing capabilities, leading to more pronounced seasonal effects, such as denser snowfall and more vibrant foliage. However, this comes at the cost of lower similarity to the reference image. Conversely, with a larger $\lambda_I$, the model prioritizes reference image fidelity, reducing its ability to perform significant edits.
Through empirical observation, we find that setting $\lambda_I$ in the range of 1 to 2.5 achieves a favorable balance between semantic editing strength and reference image preservation. This allows users to adjust $\lambda_I$ based on specific needs, trading off between edit intensity and reference consistency.

\vspace{-4mm}
\paragraph{Adaptive Attention Fusion}
To verify the effectiveness of our adaptive attention fusion in Sec.~\ref{sec:balance}, we conduct an ablation study comparing alternative fusion strategies:
\begin{itemize}
    \item Direct Replacement:$\boldsymbol{Z}^{\text{in}} =  \boldsymbol{Z}^{\prime}$, assuming full reliance on the reference-enhanced feature map.
    \item Direct Addition: $\boldsymbol{Z}^{\text{in}} = \boldsymbol{Z} + \boldsymbol{Z}^{\prime} $, treating both components as equally important.
    \item Ours: $ \boldsymbol{Z}^{\text{in}} = \boldsymbol{Z} + \text{Linear}(\boldsymbol{Z}^{\prime})$, where a learnable transformation dynamically adjusts the integration.
\end{itemize}
The results are reported in Tab.~\ref{tab:ablation}. Direct replacement performs the worst across all metrics, demonstrating that sole reliance on $\boldsymbol{Z}^{\prime}$ severely degrades image fidelity and structure preservation. Direct addition mitigates some of these issues but still lacks adaptive control, leading to suboptimal balance. Our approach, leveraging a learnable linear transformation, achieves the best equilibrium by dynamically adjusting reference influence while preserving generative capability. This validates the necessity of adaptive feature integration for high-quality image editing.

\subsection{Compatibility with Community Extensions}
\label{extensions}

A key advantage of our approach is its seamless compatibility with community-driven extensions, enabling flexible integration without modifying the base model or requiring retraining. Unlike prior methods that impose architectural constraints or demand fine-tuning, our design preserves the diffusion model’s structure while naturally incorporating external control signals.

We validate this capability with three representative models: IP-Adapter\cite{ipadapter}, ControlNet\cite{Zhangcontrolnet}, and RealCartoon3D, each serving distinct purposes—IP-Adapter and ControlNet enhance conditional control through structural and style guidance, while RealCartoon3D specializes in artistic stylization. As shown in Fig.~\ref{fig:community}, our method seamlessly integrates with each: IP-Adapter and ControlNet transform scenes into winter settings with realism and structural consistency, while RealCartoon3D produces high-quality cartoon-style conversions without artifacts.

These results underscore our model’s plug-and-play adaptability, extending diffusion-based editors without altering their core architecture. By ensuring compatibility across diverse editing paradigms, our approach enhances the model's versatility and practical applicability.




\begin{figure}[t]
    \centering
    \includegraphics[width=\linewidth]{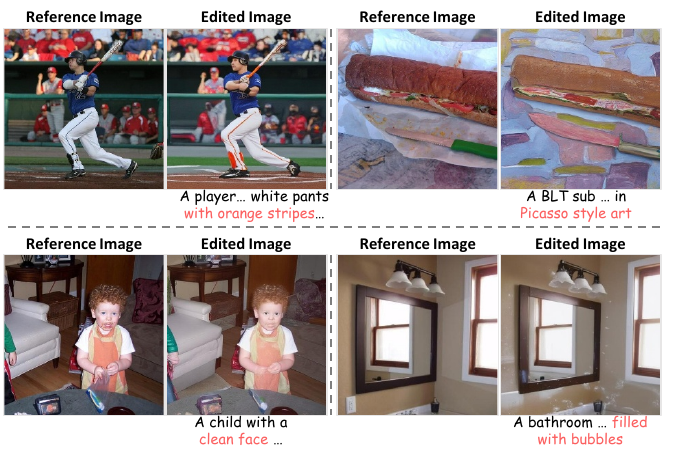}
    \vspace{-7mm}
    \caption{
    Showcases of applying DescriptiveEdit to Flux~\cite{flux}, a DiT-based text-to-image model.
    }
    \vspace{-7mm}
    \label{fig:flux}
\end{figure}

\subsection{Robustness Across Architectures}
\label{sec:flux}
To demonstrate the adaptability of our approach across different diffusion architectures, we apply it to Flux~\cite{flux}, showcasing its seamless integration with DiT-based model.
%
%
As shown in Fig.~\ref{fig:flux}, our approach consistently produces high-fidelity edits while preserving structural integrity, confirming the robustness of our adaptive feature integration strategy. By eliminating architecture-dependent constraints, our method ensures reliable performance across diverse diffusion-based frameworks, reinforcing its practicality for real-world applications.





\section{Conclusions}
\label{sec:conclusions}

We proposed a description-based method, DescriptiveEdit, that unifies semantic image editing to the text-to-image generation framework. An attention bridge enables efficient feature integration, while LoRA-based tuning ensures parameter efficiency and compatibility. Experiments show superior editing accuracy and consistency, establishing a scalable, plug-and-play paradigm for semantic image editing.

\section*{Acknowledgements}
This work was supported by Natural Science Foundation of China: No. 62406135, Natural Science Foundation of Jiangsu Province: BK20241198, and the Gusu Innovation and Entrepreneur Leading Talents: No. ZXL2024362.
This work was supported in part by the Shanghai Municipal Commission of Economy and Informatization, under Grant No. 2024-GZL-RGZN-01008.

{
    \small
    \bibliographystyle{ieeenat_fullname}
    \bibliography{main}
}

\end{document}